\documentclass[runningheads]{llncs}
\usepackage{graphicx}

\usepackage{tikz}
\usepackage{comment}
\usepackage{amsmath,amssymb} 
\usepackage{color}
\usepackage{hyperref}

\usepackage{multirow}

\usepackage[accsupp]{axessibility}  

\usepackage{cite}

\begin{document}
\pagestyle{headings}
\mainmatter
\def\ECCVSubNumber{8}  

\title{MIPI 2022 Challenge on RGBW Sensor Re-mosaic: Dataset and Report} 

\titlerunning{MIPI 2022 Challenge on RGBW Sensor Re-mosaic}
%
\author{
Qingyu Yang \and 
Guang Yang \and
Jun Jiang \and
Chongyi Li \and
Ruicheng Feng \and
Shangchen Zhou \and
Wenxiu Sun \and
Qingpeng Zhu \and
Chen Change Loy \and
Jinwei Gu \and
Lingchen Sun \and 
Rongyuan Wu \and
Qiaosi Yi \and 
Rongjian Xu \and 
Xiaohui Liu \and
Zhilu Zhang \and 
Xiaohe Wu \and 
Ruohao Wang \and 
Junyi Li \and 
Wangmeng Zuo \and 
Faming Fang
\institute{~}
\vspace{-1cm}
}
\authorrunning{Q. Yang et al.}
%

\maketitle
\let\thefootnote\relax\footnotetext{\tiny Qingyu Yang$^{1}$ (\email{yangqingyu@sensebrain.site}), Jun Jiang$^{1}$ (\email{jiangjun@sensebrain.site}),  Chongyi Li$^{4}$, Shangchen Zhou$^{4}$, Ruicheng Feng$^{4}$, Wenxiu Sun$^{2,3}$, Qingpeng Zhu$^{2}$, Chen Change Loy$^{4}$, Jinwei Gu$^{1,3}$,   are the MIPI 2022 challenge organizers ($^{1}$SenseBrain, $^{2}$SenseTime Research and Tetras.AI, $^{3}$Shanghai AI Laboratory, $^{4}$Nanyang Technological University). The other authors participated in the challenge. Please refer to Appendix~\ref{appendix:teams} for details.\\ 
\\
MIPI 2022 challenge website: \url{http://mipi-challenge.org/}
}

\begin{abstract}
Developing and integrating advanced image sensors with novel algorithms in camera systems are prevalent with the increasing demand for computational photography and imaging on mobile platforms. However, the lack of high-quality data for research and the rare opportunity for in-depth exchange of views from industry and academia constrain the development of mobile intelligent photography and imaging (MIPI). To bridge the gap, we introduce the first MIPI challenge including five tracks focusing on novel image sensors and imaging algorithms. In this paper, RGBW Joint Remosaic and Denoise, one of the five tracks, working on the interpolation of RGBW CFA to Bayer at full-resolution, is introduced. The participants were provided with a new dataset including 70 (training) and 15 (validation) scenes of high-quality RGBW and Bayer pairs. In addition, for each scene, RGBW of different noise levels was provided at 0dB, 24dB, and 42dB. All the data were captured using an RGBW sensor in both outdoor and indoor conditions. The final results are evaluated using objective metrics including PSNR, SSIM~\cite{ssim}, LPIPS~\cite{lpips}, and KLD. A detailed description of all models developed in this challenge is provided in this paper. More details of this challenge and the link to the dataset can be found at \href{https://github.com/mipi-challenge/MIPI2022}{https://github.com/mipi-challenge/MIPI2022}

\keywords{RGBW, Remosaic, Bayer, Denoise, MIPI challenge}
\end{abstract}

\section{Introduction}
\begin{figure}[!ht]
\centering
\includegraphics[width=0.5\textwidth]{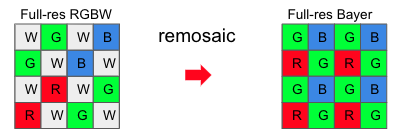}
\caption{The RGBW remosaic task.}
\label{fig:quad}
\setlength{\belowcaptionskip}{0pt plus 3pt minus 2pt}
\end{figure}

RGBW is a new type of CFA pattern (Fig.~\ref{fig:quad}) designed for image quality enhancement under low light conditions. Thanks to the higher optical transmittance of white pixels over conventional red, green, and blue pixels, the signal-to-noise ratio (SNR) of the sensor output becomes significantly improved, thus boosting the image quality especially under low light conditions. Recently several phone OEMs, including Transsion, Vivo, and Oppo have adopted RGBW sensors in their flagship smartphones to improve the camera image quality~\cite{oppoRGBW ,  vivoRGBW, TranssionRGBW}.

On the other hand, conventional camera ISPs can only work with Bayer patterns, thereby requiring an interpolation procedure to convert RGBW to a Bayer pattern. The interpolation process is usually referred to as remosaic, and a good remosaic algorithm should be able (1) to get a Bayer output from RGBW with least artifacts, and (2) to fully take advantage of the SNR and resolution benefit of white pixels.

The remosaic problem becomes more challenging when the input RGBW becomes noisy, especially under low light conditions. A joint remosaic and denoise task is thus in demand for real world applications.

\begin{figure}[!ht]
\centering
\includegraphics[width=\textwidth]{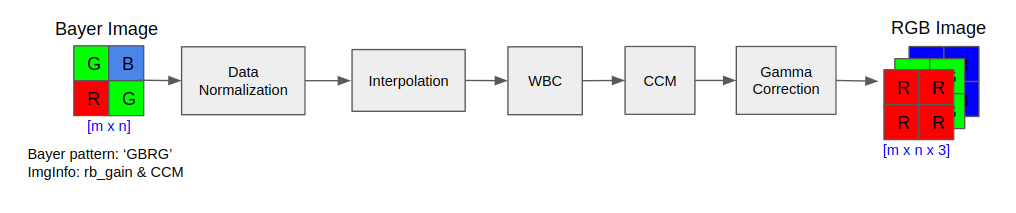}
\caption{An ISP to visuailize the output Bayer and to calculate the loss function.}
\label{fig:simple_isp}
\setlength{\belowcaptionskip}{0pt plus 3pt minus 2pt}
\end{figure}

In this challenge, we intend to remosaic the RGBW input to obtain a Bayer at the same spatial resolution. The solution is not necessarily deep-learning. However, to facilitate the deep learning training, we provide a dataset of high-quality RGBW and Bayer pairs, including 100 scenes (70 scenes for training, 15 for validation, and 15 for testing). We provide a Data Loader to read these files and show a simple ISP in Fig.~\ref{fig:simple_isp} to visualize the RGB output from the Bayer and to calculate loss functions. The participants are also allowed to use other public-domain datasets. The algorithm performance is evaluated and ranked using objective metrics: Peak Signal-to-Noise Ratio (PSNR), Structural Similarity Index (SSIM)~\cite{ssim}, Learned Perceptual Image Patch Similarity (LPIPS)~\cite{lpips}, and KL-divergence (KLD). The objective metrics of a baseline method are available as well to provide a benchmark. 

This challenge is a part of the Mobile Intelligent Photography and Imaging (MIPI) 2022 workshop and challenges emphasizing the integration of novel image sensors and imaging algorithms, which is held in conjunction with ECCV 2022. It consists of five competition tracks:
\begin{enumerate}
  \item RGB+ToF Depth Completion uses sparse and noisy ToF depth measurements with RGB images to obtain a complete depth map.
  \item Quad-Bayer Re-mosaic converts Quad-Bayer RAW data into Bayer format so that it can be processed by standard ISPs.
  \item RGBW Sensor Re-mosaic converts RGBW RAW data into Bayer format so that it can be processed by standard ISPs.
  \item RGBW Sensor Fusion fuses Bayer data and a monochrome channel data into Bayer format to increase SNR and spatial resolution.
  \item Under-display Camera Image Restoration improves the visual quality of the image captured by a new imaging system equipped with an under-display camera.
\end{enumerate}

\section{Challenge}
To develop high-quality RGBW Remosaic solution, we provide the following resources for participants:
\begin{itemize}
    \item A high-quality RGBW and Bayer dataset: As far as we know, this is the first and only dataset consisting of aligned RGBW and Bayer pairs, relieving the pain of data collection to develop learning-based remosaic algorithms;
    \item A data processing code with Data Loader to help participants get familiar with the provided dataset;
    \item A simple ISP including basic ISP blocks to visualize the algorithm output and to calculate the loss function on RGB results;
    \item A set of objective image quality metrics to measure the performance of a developed solution.
\end{itemize}

\subsection{Problem Definition}
RGBW remosaic aims to interpolate the input RGBW CFA pattern to obtain a Bayer of the same resolution. The remosaic task is needed mainly because current camera ISPs usually cannot process CFAs other than the Bayer pattern. In addition, the remosaic task becomes more challenging when the noise level gets higher, thus requiring more advanced algorithms to avoid image quality artifacts. In addition to the image quality requirement, RGBW sensors are widely used in smartphones with limited computational budget and battery life, thus requiring the remosaic algorithm to be lightweight at the same time. While we do not rank solutions based on running time or memory footprint, computational cost is one of the most important criteria in real applications.

\subsection{Dataset: Tetras-RGBW-RMSC}

\begin{figure}[!ht]
\centering
\includegraphics[width=\textwidth]{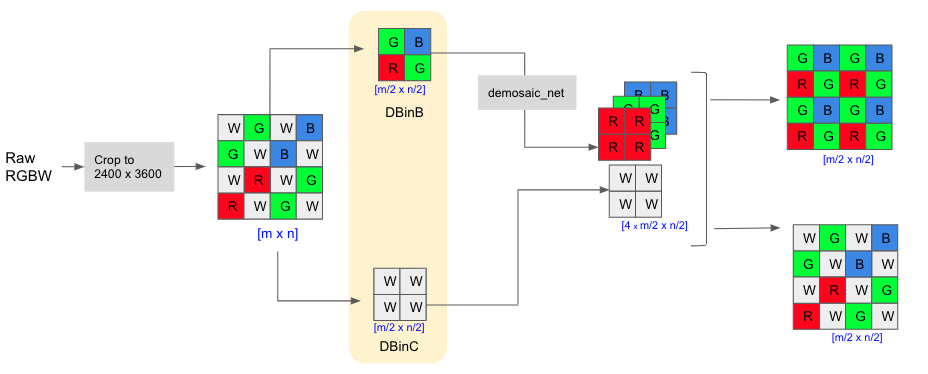}
\caption{Data generation of the RGBW remosaic task. The RGBW raw data is captured using a RGBW sensor and cropped to be a size of $2400\times3600$. A Bayer (DBinB) and white (DBinC) image are obtained by averaging the same color in the diagonal direction within a $2\times2$ block. We demosaic the Bayer (DBinB) to get an RGB using the DemosaicNet~\cite{gharbi2016deep}. The white (DBinC) is concatenated to the RGB image to have RGBW for each pixel, which is in turn mosaiced to get the input RGBW and aligned ground truth Bayer.}
\label{fig:data_gen}
\setlength{\belowcaptionskip}{0pt plus 3pt minus 2pt}
\end{figure}

The training data contains 70 scenes of aligned RGBW (input) and Bayer (ground truth) pairs. For each scene, noise is sythesized on the 0dB RGBW input to provide the noisy RGBW input at 24dB and 42dB respectively. The synthesized noise consists of read noise and shot noise, and the noise models are measured on an RGBW sensor. The data generation steps are shown in Fig.~\ref{fig:data_gen}. The testing data contains RGBW input of 15 scenes at 0dB, 24dB, and 42dB, but the ground truth Bayer results are not available to participants. 

\subsection{Challenge Phases}
The challenge consisted of the following phases:
\begin{enumerate}
    \item Development: The registered participants get access to the data and baseline code, and are able to train the models and evaluate their run-time locally.
    \item Validation: The participants can upload their models to the remote server to check the fidelity scores on the validation dataset, and to compare their results on the validation leaderboard.
    \item Testing: The participants submit their final results, codes, models, and factsheets.
\end{enumerate}

\subsection{Scoring System}
\subsubsection{Objective Evaluation}
The evaluation consists of (1) the comparison of the remosaic output (Bayer) with the reference ground truth Bayer, and (2) the comparison of RGB from the predicted and ground truth Bayer using a simple ISP (the code of the simple ISP is provided). We use
\begin{enumerate}
    \item Peak Signal-to-Noise Ratio (PSNR)
    \item Structural Similarity Index Measure (SSIM)~\cite{ssim}
    \item Kullback–Leibler Divergence (KLD)
    \item Learned Perceptual Image Patch Similarity (LPIPS)~\cite{lpips}
\end{enumerate}
to evaluate the remosiac performance. The PSNR, SSIM, and LPIPS will be applied to the RGB from the Bayer using the provided simple ISP code, while KLD is evaluated on the predicted Bayer directly.

A metric weighting PSNR, SSIM, KLD, and LPIPS is used to give the final ranking of each method, and we will report each metric separately as well. The code to calculate the metrics is provided. The weighted metric is shown below. The M4 score is between 0 and 100, and the higher score indicates the better overall image quality.

\begin{equation}
    \text{M4} = PSNR \cdot \text{SSIM}  \cdot 2^{1-\text{LPIPS}-\text{KLD}} .
\label{eq:M4}
\end{equation}
For each dataset we report the average results over all the processed images belonging to it.

\section{Challenge Results}

The results of the top three teams are shown in Table.~\ref{tab:results}. In the final test phase, we verified their submission using their code. \textbf{op-summer-po}, \textbf{HIT-IIL}, and \textbf{Eating, Drinking, and Playing} are the top three teams ranked by M4 as presented in Eq.~\eqref{eq:M4}, and \textbf{op-summer-po} shows the best overall performance. The proposed methods are described in Section \ref{sec:methods} and the team members and affiliations are listed in Appendix \ref{appendix:teams}.

\begin{table}[!ht]  
    \centering
    
    \begin{tabular}{l | llll | l}
    \hline
        \textbf{Team name} & \textbf{PSNR} & \textbf{SSIM} & \textbf{LPIPS} & \textbf{KLD} &  \textbf{M4}\\ \hline  \hline
        \text{op-summer-po}                                & 36.83         & 0.957             & 0.115      & 0.018         & 64.89 \\ \hline
        \text{HIT-IIL}                                              & 36.34         & 0.95               & 0.129      & 0.02           & 63.12 \\ \hline
        \text{Eating, Drinking, and Playing}         & 36.77         & 0.957              & 0.132      & 0.019         & 63.98 \\ \hline
    \end{tabular}
    \caption{MIPI 2022 Joint RGBW Remosaic and Denoise challenge results and final rankings. PSNR, SSIM, LPIPS, and KLD are calculated between the submitted results from each team and the ground truth data. A weighted metric, M4, by Eq.~\eqref{eq:M4} is used to rank the algorithm performance, and the top three teams with the highest M4 are included in the table.  
    \label{tab:results}}
\end{table}

\begin{figure}[!ht]
\setlength{\abovecaptionskip}{0.cm}
\setlength{\belowcaptionskip}{-0.cm}
\centering
\includegraphics[width=\textwidth]{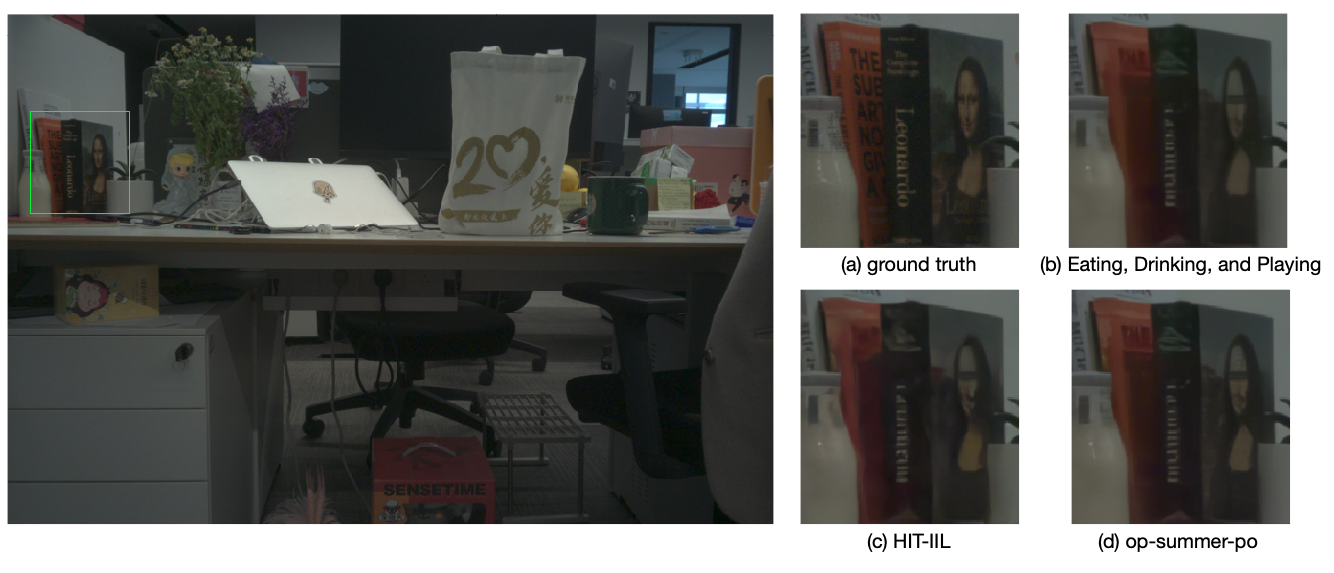}
\caption{Qualitative image quality (IQ) comparison. The results of one of the test scenes (42dB) are shown. While the top three remosaic methods achieve high objective IQ metrics in Table.~\ref{tab:results}, details and texture loss are noticeable on the book cover. The texts on the book are barely interpretable in (b), (c) and (d). The RGB images are obtained by using the ISP in Fig.~\ref{fig:simple_isp}, and its code is provided to participants.}
\label{fig:IQ2}
\end{figure}

\begin{figure}[!ht]
\setlength{\abovecaptionskip}{0.cm}
\setlength{\belowcaptionskip}{-0.cm}
\centering
\includegraphics[width=\textwidth]{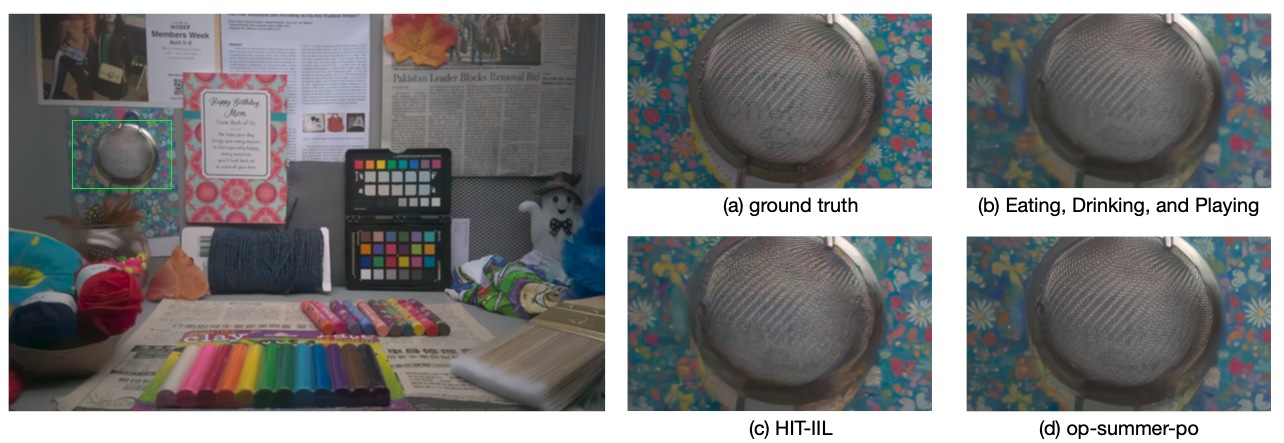}
\caption{Qualitative image quality (IQ) comparison. The results of one of the test scenes (42dB) are shown. Oversmoothing in the top three methods in Table.~\ref{tab:results} can be found when compared with the ground truth. The text under the mesh can barely be recognized, and the mesh texture become distorted in (b),(c) and (d). The RGB images are obtained by using the ISP in Fig.~\ref{fig:simple_isp}, and its code is provided to participants.}
\label{fig:IQ1}
\end{figure}

To learn more about the algorithm performance, we evaluated the qualitative image quality in addition to the objective IQ metrics in Fig.~\ref{fig:IQ2} and Fig.~\ref{fig:IQ1} respectively. While all teams in Table.~\ref{tab:results} have achieved high PSNR and SSIM, the detail and texture loss can be found on the book cover in Fig.~\ref{fig:IQ2} and on the mesh in Fig.~\ref{fig:IQ1}. When the input has a large amount of noise, oversmoothing tends to yield higher PSNR at the cost of detail loss perceptually.  

In addition to benchmarking the image quality of remosaic algorithms, computational efficiency is evaluated because of wide adoptions of RGBW sensors on smartphones. We measured the runnnig time of the remosaic solutions of the top three teams (based on M4 by Eq.~\eqref{eq:M4}) in Table.~\ref{tab:runtime}. While running time is not employed in the challenge to rank remosaic algorithms, the computational cost is of critical importance when developing algorithms for smartphones. HIT-IIL achieved the shortest running time among the top three solutions on a workstation GPU (NVIDIA Tesla V100-SXM2-32GB). With sensor resolution of mainstream smartphones reaching 64M or even higher, power-efficient remosaic algorithms are highly desirable.

\begin{table}[!ht]  
    \centering
    
    \begin{tabular}{l | l | l}
    \hline
        \textbf{Team name} & \textbf{1200$\times$1800 (measured)}  &  \textbf{64M} (estimated)\\ \hline  \hline
        \text{op-summer-po}                                      & 6.2s                   &  184s \\ \hline
        \text{HIT-IIL}                                                    &  \textbf{4.1s}     &  \textbf{121.5s} \\ \hline
        \text{Eating, Drinking and Playing}                & 10.4s                  &  308s  \\ \hline

    \end{tabular}
    \caption{Running time of the top three solutions ranked by Eq.~\eqref{eq:M4} in the 2022 Joint RGBW Remosaic and Denoise challenge. The running time of input of $1200\times1800$ was measured, while the running time of a 64M input RGBW was based on the estimation. The measurement was taken on an NVIDIA Tesla V100-SXM2-32GB GPU.
    \label{tab:runtime}}
\end{table}

\section{Challenge Methods}
\label{sec:methods}
In this section, we describe the solutions submitted by all teams participating in the final stage of MIPI 2022 Joint RGBW Remosaic and Denoise Challenge. 

\subsection{op-summer-po}
\begin{figure}[!ht]
\setlength{\abovecaptionskip}{0.cm}
\setlength{\belowcaptionskip}{-0.cm}
\centering
\includegraphics[width=0.9\textwidth]{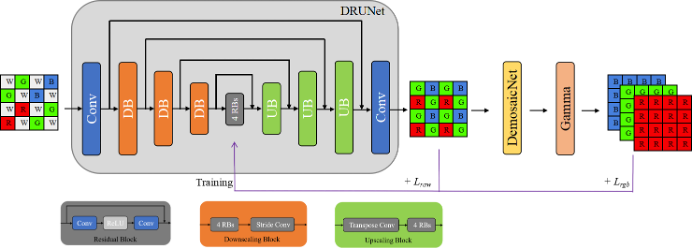}
\caption{The model architecture of op-summer-po.}
\label{fig:nn-op-summer-po}
\end{figure}
op-summer-po proposed a framework based on DRUNet~\cite{zhang2021plug} as shown in Fig.~\ref{fig:nn-op-summer-po}, which has four scales with 64, 128, 256, and 512 channels respectively. Each scale has an identity skip connection between the stride convolution (stride = 2) and the transpose convolution. This connection concatenates encoder and decoder features. Each encoder or decoder includes four residual blocks.

The input of the framework is the raw RGBW image, and the output is the estimated raw Bayer image. Then the estimated raw Bayer image is sent to DemosaicNet~\cite{gharbi2016deep} and Gamma Transform to get the full-resolution RGB image. Moreover, two LPIPS~\cite{lpips} functions in both raw and RGB domains are used to get a better perception quality.

\subsection{HIT-IIL}
\begin{figure}[!ht]
\setlength{\abovecaptionskip}{0.cm}
\setlength{\belowcaptionskip}{-0.cm}
\centering
\includegraphics[width=0.99\textwidth]{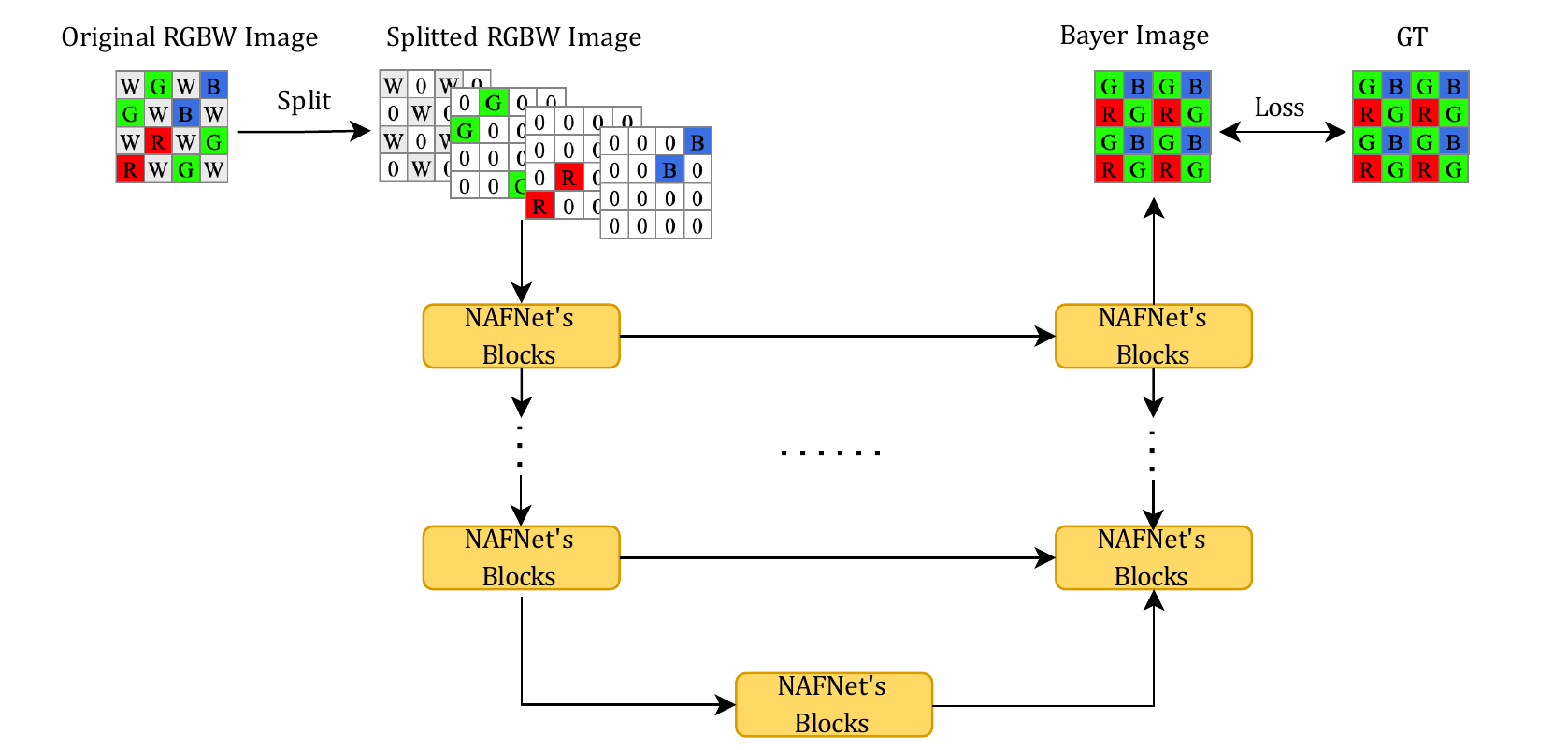}
\caption{The model architecture of HIT-IIL.}
\label{fig:nn-HIT-IIL}
\end{figure}
\text{HIT-IIL} proposed an end-to-end method to jointly learn RGBW re-mosaicing and denoising. They split RGBW images into 4-channel while maintaining image size, as shown in Fig.~\ref{fig:nn-HIT-IIL}. Specifically, they repeated the RGBW image as 4 channels, then made each channel represent a color type (i.e., one of white, green, blue, and red). When the pixel color type is different from the color represented by the channel, its value is set to 0. Such input mode provides the positional information of the color for the network and significantly improves performance in experiments.

As for the end-to-end network, they adopt NAFNet~\cite{chen2022simple} to re-mosaic and denoise RGBW images. NAFNet contains the 4-level encoder-decoder and bottleneck. For the encoder, the numbers of NAFNet’s blocks for each level are 2, 4, 8, and 24. For the decoder, the numbers of NAFNet’s blocks for the 4 levels are all 2. In addition, the number of NAFNet’s blocks for the bottleneck is 12.

\subsection{Eating, Drinking and Playing}
\begin{figure}[!ht]
\setlength{\abovecaptionskip}{0.cm}
\setlength{\belowcaptionskip}{-0.cm}
\centering
\includegraphics[width=0.99\textwidth]{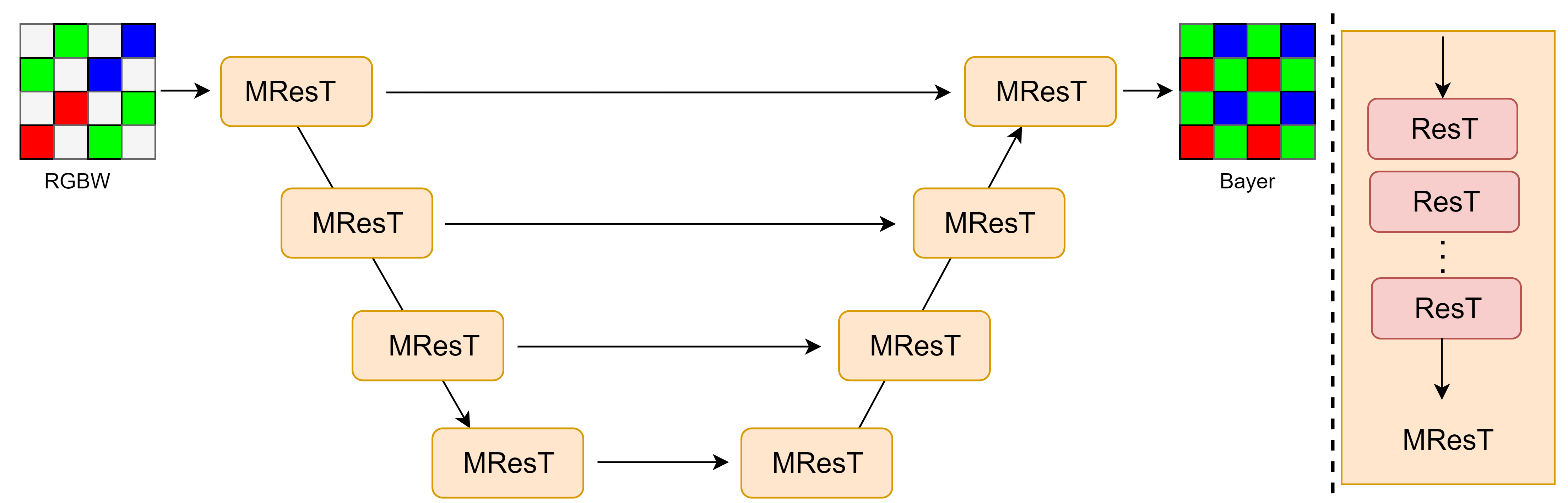}
\caption{The model architecture of Eating, Drinking and Playing.}
\label{fig:nn-eating}
\end{figure}
Eating, Drinking and Playing proposed a UNet-Transformer for RGBW joint remosaic and denoise, and its overall architecture is in Fig.~\ref{fig:nn-eating}. To be more specific, they aimed to estimate a raw Bayer image $O$ from the corresponding raw RGBW image $I$ . To achieve this, they decreased the distance between the output $O$ and the ground truth raw Bayer image by minimizing $L1$ loss. 

\begin{figure}[!ht]
\setlength{\abovecaptionskip}{0.cm}
\setlength{\belowcaptionskip}{-0.cm}
\centering
\includegraphics[width=0.69\textwidth]{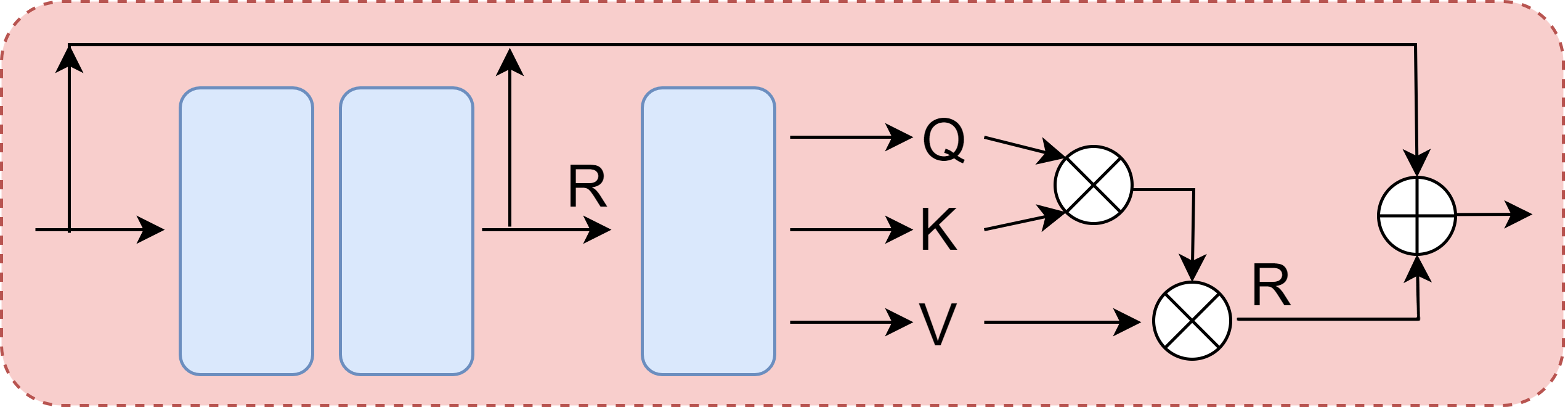}
\caption{The structure of the Res-Transformer (ResT) block. R is the reshape and the blue block is the convolution layer.}
\label{fig:nn-eating2}
\end{figure}

Similar to DRUNet~\cite{zhang2021plug}, the UNet-Transformer adopts an encoder-decoder structure and has four levels with 64, 128, 256, and 512 channels respectively. The main difference between UNet-Transformer and DRUNet is that UNet-Transformer adopts Multi-ResTransformer (MResT) blocks rather than residual convolution blocks in each level of the encoder and decoder. As the main component of UNet-Transformer, the MResT cascades multiple Res-Transformer (ResT) blocks, which can study local and global information simultaneously. The structure of ResT is illustrated in Fig.~\ref{fig:nn-eating2}. In the ResT, the first two convolution layers were adopted to study the local information, and then they employed a self-attention to learn the global information.

\section{Conclusions}
In this paper, we summarized the Joint RGBW Remosaic and Denoise challenge in the first Mobile Intelligent Photography and Imaging workshop (MIPI 2022) held in conjunction with ECCV 2022. The participants were provided with a high-quality training/testing dataset for RGBW remosaic and denoise, which is now available for researchers to download for future research. We are excited to see so many submissions within such a short period, and we look forward for more research in this area.

\section{Acknowledgements}
We thank Shanghai Artificial Intelligence Laboratory, Sony, and Nanyang Technological University to sponsor this MIPI 2022 challenge. We thank all the organizers and participants for their great work. 

\appendix
\section{Teams and Affiliations}
\label{appendix:teams}
\tiny
\textbf{op-summer-po} \\
\textbf{Title}: Two LPIPS Functions in Raw and RGB domains for RGBW Joint Remosaic and Denoise \\
\textbf{Members}:
$^{1}$Lingchen Sun, (slcbbd111@sina.com), $^{1}$Rongyuan Wu,  $^{1}$,$^{2}$Qiaosi Yi \\
\textbf{Affiliations}:
$^{1}$OPPO Research Institute, $^{2}$East China Normal University\\
\\
\\
\textbf{HIT-IIL} \\
\textbf{Title}: Data Input and Augmentation Strategies for RGBW Image Re-Mosaicing\\
\textbf{Members}:
Rongjian Xu (ronjon.xu@gmail.com), 
Xiaohui Liu, Zhilu Zhang, Xiaohe Wu, Ruohao Wang, Junyi Li, Wangmeng Zuo \\
\textbf{Affiliations}:
Harbin Institute of Technology\\
\\
\\
\textbf{Eating, Drinking, and Playing} \\
\textbf{Title}: UNet-Transformer for RGBW Joint Remosaic and Denoise \\
\textbf{Members}:
$^{1}$,$^{2}$Qiaosi Yi (51205901027@stu.ecnu.edu.cn), $^{1}$Rongyuan Wu, $^{1}$Lingchen Sun, $^{2}$Faming Fang \\
\textbf{Affiliations}:
$^{1}$OPPO Research Institute, $^{2}$East China Normal University\\
\\
\\

%
%
\bibliographystyle{splncs04}
\bibliography{egbib}
\end{document}